# Metrics for Explainable AI: Challenges and Prospects

Robert R. Hoffman
Institute for Human and Machine Cognition [rhoffman@ihmc.us]
Shane T. Mueller
Michigan Technological University [shanem@mtu.edu]
Gary Klein
Macrocognition, LLC [gary@macrocognition.com]
Jordan Litman
Institute for Human and Machine Cognition [jlitman@ihmc.us]

**Abstract**
The question addressed in this paper is: If we present to a user an AI system that explains how it works, how do we know whether the explanation works and the user has achieved a pragmatic understanding of the AI? In other words, how do we know that an explanainable AI system (XAI) is any good? Our focus is on the key concepts of measurement. We discuss specific methods for evaluating: (1) the goodness of explanations, (2) whether users are satisfied by explanations, (3) how well users understand the AI systems, (4) how curiosity motivates the search for explanations, (5) whether the user's trust and reliance on the AI are appropriate, and finally, (6) how the human-XAI work system performs. The recommendations we present derive from our integration of extensive research literatures and our own psychometric evaluations.

# 1 Introduction

For decision makers who rely upon Artificial Intelligence (AI), analytics and data science, explainability is an issue. If a computational system relies on a simple statistical model, decision makers can understand it and convince executives who have to sign off on a system that it is reasonable and that seems fair. They can justify the analytical results to shareholders, regulators, etc. But for Machine Learning and Deep Net systems, they can no longer do this. There is a need for ways to explain the computational system to the decision maker so that they know that their full process is going to be reasonable.

> ... current efforts face unprecedented difficulties: contemporary models are more complex and less interpretable than ever; [AI is] used for a wider array of tasks, and are more pervasive in everyday life than in the past; and [AI is] increasingly allowed to make (and take) more autonomous decisions (and actions). Justifying these decisions will only become more crucial, and there is little doubt that this field will continue to rise in prominence and produce exciting and much needed work in the future (Biran and Cotton, 2017, p. 4).

This quotation brings into relief the importance of "Explainable AI" (XAI.) A proposed regulation before the European Union (Goodman and Flaxman, 2016) prohibits "automatic processing" unless user's rights are safeguarded. Users have a "right to an explanation" concerning algorithm-created decisions that are based on personal information. Future laws may restrict AI, which represents a challenge to industry.

Expert Systems researchers have previously implemented methods for explanation (Clancey, 1984, 1986; McKeown & Swartout, 1987; Moore & Swartout, 1990). In a sense, explanation is what and Intelligent Tutoring Systems were (and are) all about (Forbus & Feltovich, 2001; Poulson & Richardson, 1988; Psotka, Massey & Mutter, 1988; Ritter & Feurzeig, 1988; Sleeman & Brown, 1982). AI systems are receiving considerable attention in the recent popular press (Alang, 2017; Bornstein, 2016; Champlin, Bell & Schocken, 2017; Harford, 2014; Hawkins, 2017; Kuang, 2017; Pavlus, 2017; Pinker, 2017; Schwiep, 2017; Voosen, 2017; Weinberger, 2017). Reporting and opinion pieces have discussed social justice, equity, and fairness issues that are implicated by AI (e.g., Felten, 2017).

The goals of explanation involve answering questions such as, "How does it work?" and "What mistakes can it make?" and "Why did it just do that?" The question addressed in this paper is: If we present to a user an AI system that explains how it works, how do we go about measuring whether or not it works, whether it works well, and whether the user has achieved a pragmatic understanding? Our focus in this paper is on the key concepts of measurement for the evaluation of XAI systems and human-machine performance.

## 1.1 Key Measurement Concepts

The concept or process of explanation or understanding has been explored in one way or another by scholars and scientists of all schools and specializations, spanning all of human civilization. To say that the pertinent literature is enormous is an understatement. In modern times, the concept is a focus in Philosophy of Science, Psychology (Cognitive, Developmental, Social, Organizational), Education and Training, Team Science, and Human Factors.

> While explainable AI is only now gaining widespread visibility, [there is a] continuous history of work on explanation and can provide a pool of ideas for researchers currently tackling the task of explanation (Biran and Cotton, 2017, p. 4).

Key concepts include causal reasoning and abductive inference, comprehension of complex systems, counterfactual reasoning, and contrastive reasoning. For reviews of the literature, see Miller (2017) and Hoffman, Klein and Mueller (2018).

A conceptual model of the XAI explaining process is presented in Figure 1. This diagram highlights four major classes of measures. Initial instruction in how to use an AI system will enable the user to form an initial mental model of the task and the AI system. Subsequent experience, which can include system-generated explanations, would enable to participant to refine their mental model, which should lead to better performance and appropriate trust and reliance.

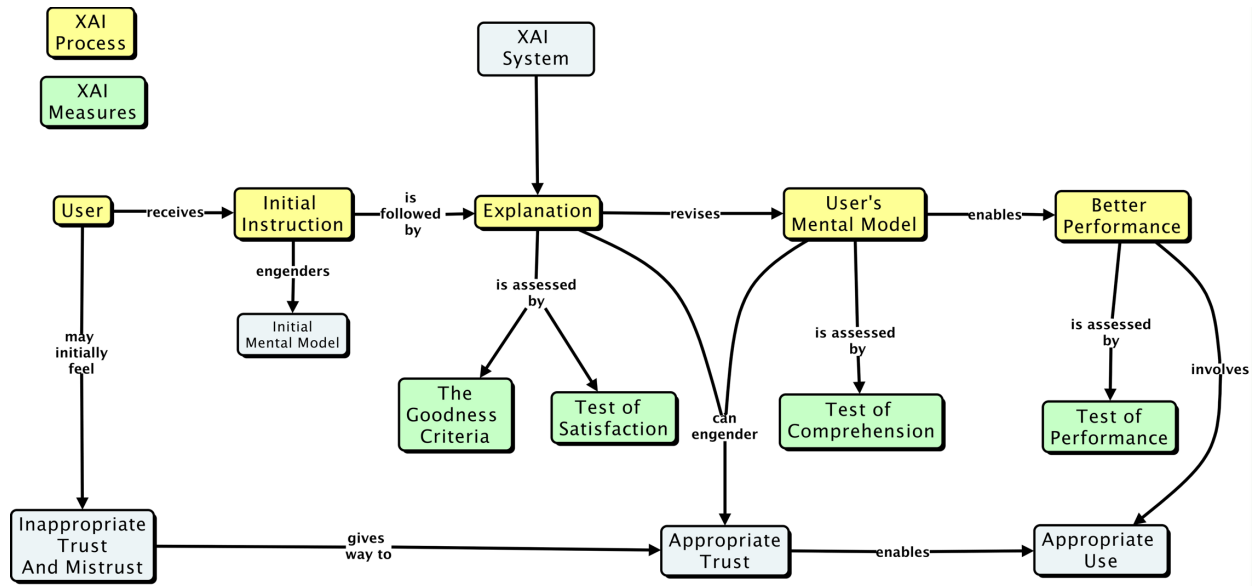

Figure 1. A conceptual model of the process of explaining, in the XAI context.

By hypothesis, explanations that are good and are satisfying to users enable users to develop a good mental model. In turn, their good mental model will enable them to develop appropriate trust in the AI and perform well when using the AI. To evaluate this model of the explanation process, a number of types of measures are required (Miller, 2017). In the remainder of this report we detail each of the four classes of measures and offer specific methodological suggestions.

# 2 Explanation Goodness and Satisfaction

The property of "being an explanation" is not a property of statements, it is an interaction. What counts as an explanation depends on what the learner/user needs, what knowledge the user (learner) already has, and especially the user's goals. This leads to a consideration of function and context of the AI system (software, algorithm, tool), That is, why does a given user need an explanation? In the various pertinent literatures, this is expressed in terms of the different kinds of questions that a user might have. These "triggers" for explanation are listed in Table 1.

Table 1.  Triggers and goals.

| TRIGGERS | USER/LEARNER'S GOAL |
|---|---|
| How do I use it? | Achieve the primary ask goals |
| How does it work? | Feeling of satisfaction at having achieved an understanding of the system, in general (global understanding) |
| What did it just do? | Feeling of satisfaction at having achieved a understanding of how the system made a particular decision (local understanding) |
| What does it achieve? | Understanding of the system's functions and uses |
| What will it do next? | Feeling of trust based on the observability and predictability of the system |
| How much effort will this take? | Feeling of effectiveness and achievement of the primary task goals |
| What do I do if it gets it wrong? | Desire to avoid mistakes |
| How do I avoid the failure modes? | Desire to mitigate errors |
| What would it have done if x were different? | Resolution of curiosity at having achieved an understanding of the system |
| Why didn't it do z? | Resolution of curiosity at having achieved an understanding of the local decision |

Thus, the seeking of an explanation can tacitly be an expression of a need for a certain kind of explanation, to satisfy certain user purposes of user goals.

## 2.1 Explanation Goodness

Looking across the scholastic and research literatures on explanation, we find assertions about what makes for a good explanation, from the standpoint of statements as explanations,. There is a general consensus on this; factors such as clarity and precision. Thus, one can look at a given explanation and make an a priori (or decontextualized) judgment as to whether or not it is "good." Appendix A presents a Goodness Checklist that can be used by XAI researchers to either try and design goodness into the explanations that their XAI system generates, or to evaluate the a priori goodness of the explanations that an XAI system generates. In a proper experiment, the researchers who complete the checklist, with reference to some particular AI-generated explanation, would not be the ones who created the XAI system under study. Using the Goodness checklist, those independent judges ask, Are the researchers right in claiming that their explanations are good?

## 2.2 Explanation Satisfaction

While an explanation might be deemed good in the manner described above, it may at the same time not be adequate or satisfying to users-in-context. Explanation Satisfaction is defined as the degree to which users feel that they understand the AI system or process being explained to them. Compared to Goodness, as we defined it above, satisfaction is a contextualized, a posteriori judgment of explanations.

Based on our review of the psychological literature on explanation, including theoretical and empirical work by Muir (1987, 1994) and by Cahour and Forzy (2009), we identified several key attributes of explanations: understandability, feeling of satisfaction, sufficiency of detail, completeness, usefulness, accuracy, and trustworthiness. These terms were incorporated into an initial pool of Likert scale items that we constructed for review and evaluation.

Standard practice in psychometrics involves assuring that a scale is internally consistent or reliable (evaluated primarily by the method of Cronbach's alpha, which is based on inter-item covariance). For the XAI context, it is immediate importance to determine the validity of an Explanation Satisfaction scale.

## 2.3 Scale Validation: Content Validity

Psychometric theory describes several methods for evaluating validity, all of which ultimately refer to the question, Does the scale measure what it is intended to measure? Two distinct kinds of validity are immediately pertinent to the development of a scale of Explanation Satisfaction. These are described in Table 2.

Table 2. Types of validity that are immediately pertinent to an evaluation of the Satisfaction Scale.

| Surface (i.e., "face") Validity |
|---|
| This is a judgment that the items appear to measure what they are intended to measure because they refer literally and explicitly to the conceptual measurables that the instrument is supposed to measure. For example, a questionnaire item that asks you to rate how smart you think you are compared to the average person would have surface validity as a measure of self-perceived intelligence since it refers explicitly to the concept of intelligence, and asks for a rating that treats intelligence as a scalar variable. (Setting aside for now the fact that the rating might be biased.) |
| Construct Validity |

> Construct validity is the analysis of the test in terms of some theoretical framework, or the reliance on an accepted theoretical framework in the construction of the test and the individual items. "[It] reflects the degree to which the measurement instrument spans the domain of the construct's theoretical definition; it is the extent to which a measurement instrument captures the different facets of a construct" (Rungtusanatham, 1998, p.10). Thus, the measurements fall on a measurement scale that is interpreted as a measure of the theoretical concept (Hoffman, 2010).

These aspects of validity can be evaluated by having domain experts make judgments as to whether the scale items are meaningful and measurable indicators of the relevant psychological construct, and consistent with the accepted theoretical framework for the construct. We conducted a validation study with the cooperation of participants and attendees at a recent DARPA-sponsored meeting of XAI researchers. The participants could express their judgments by rating the degree to which each scale item is essential to measuring the construct.

Participants were asked to evaluate the scale using the Content Validity Ratio (CVR) method (Lawshe, 1975). The CVR is useful for quantitatively assessing the strength of each scale item with a small group of raters. The scale that was presented to participants is presented in Table 3.

Table 3. The CVR version of the Explanation Satisfaction scale.

| For each item listed below, please indicate whether you believe that the item is: <br> • Essential for measuring Explanation Satisfaction, <br> • Useful but not Essential, or <br> • Not Necessary for measuring Explanation Satisfaction. <br><br> Indicate your response by circling the appropriate number. | | Essential | Useful but not essential | Not Necessary |
|---|---|---|---|---|
| 1. | I understand this explanation of how the [software, algorithm, tool] works. | 1 | 2 | 3 |
| 2. | This explanation of how the [software, algorithm, tool] works is satisfying. | 1 | 2 | 3 |
| 3. | This explanation of how the [software, algorithm, tool] works has sufficient detail. | 1 | 2 | 3 |
| 4. | This explanation of how the [software, algorithm, tool] works contains irrelevant details. | 1 | 2 | 3 |
| 5. | This explanation of how the [software, algorithm, tool] works seems complete. | 1 | 2 | 3 |
| 6. | This explanation of how the [software, algorithm, tool] works tells me how to use it. | 1 | 2 | 3 |
| 7. | This explanation of how the [software, algorithm, tool] works is useful to my goals. | 1 | 2 | 3 |

| 8. | This explanation of says how accurate the [software, algorithm, tool] is. | 1 | 2 | 3 |
|---|---|---|---|---|
| 9. | This explanation lets me judge when I should trust and not trust the [software, algorithm, tool] | 1 | 2 | 3 |

In order to get reasonably stable psychometric estimates for evaluating the items' communality, a rule of thumb (in the psychometric literature) is that one wants 5-10 respondents. At the DARPA PI meeting we received responses from 35 domain practitioners.

Six of the participants identified as software engineers, four self-identified as graduate students (primarily in computer science or engineering), one as a test pilot, and three did not specify their profession. Fifteen individuals self-identified as computer scientists (Ph.D. and BS/MS degrees). Three participants self-identified as psychologists. The remaining 21 participants self-identified as researchers, professors, or research managers (fields including computer science, engineering, psychology, and mathematics).

In keeping with recommendations on using the CVR method for item selection (Davis, 1992; Gilbert & Prion, 2016; Lynn, 1986), items that are rated as either "essential" or "useful" by more than 50% of the professionals were selected for further analysis; items that did not meet this criteria were dropped from further consideration. For each selected item, the CVR was calculated by dividing the number of respondents who rated the item as "essential" or "useful" to the total number of participants. CVR ratios are calculated for each item separately, by dividing the number of raters who rated an item as either "essential" or "useful" to the total number of participants. The resulting ratio values for all scale items are aggregated and transformed to a scale that ranges between −1 (perfect disagreement) and +1 (perfect agreement), with CVR values above zero indicating that over half of our domain professionals judged the item to be a highly valid indicator of ES as a psychological construct.

The average CVR for all participants who self-identified as having Ph.D. degrees was 0.43. The CVR for the computer scientists was 0.60. The average CVR for the self-identified "Researchers" was 0.43, and for "Others" was 0.57. According to Ayre and Scally (2014) the Critical Value of the average CVR should be 0.50 or greater. Given our relatively large sample size, we take our results (positive CVRs in the range of 0.43 to 0.60) to indicate reasonably high agreement as to the validity of the scale items.

Many participants reported that they found item number 4 to be confusing. This is the only item that would be reverse scored (i.e., an explanation would be unsatisfying if it contained irrelevant details). For this reason, we recalculated the Critical Values after dropping this scale item. All other items were rated as either "Essential" or "Useful" measures of Explanation Satisfaction. It is interesting to note the Computer Scientists and Electrical Engineers collectively showed higher agreement than the Psychologists. This seems to be due to the high ratings given to scale item number 7 by the individuals who self-identified as psychologists. They appear to have been emphasizing the importance of the goal relevance of explanations.

We assessed the internal consistency of the ES scale. Cronbach's alpha was 0.86; ES items had item-total correlations ranging from 0.41 to 0.76 and an average inter-item correlation of 0.71, indicative of excellent internal consistency reliability, especially for such a brief scale that is very easy to administer and score (Comrey, 1988; Cortina, 1993; Streiner, 2003).

## 2.4 Scale Validation: Discriminant Validity

Our next step was to test the discriminant validity of the ES scale, that is, does it differentiate between relatively good and poor explanations. The scale resulting from the CVR analysis was subjected to analysis of its discriminant validity using three test cases (How does a cruise control work? How do computers predict hurricanes? How do cell phones find directions?). For each test case, a small group of experts were employed to generate what they felt were "good" explanations, and the researchers themselves developed what they felt were "bad" versions. The "bad" versions were sparse, had irrelevant details, and some incorrect statements. The materials are presented in Appendix B.

Eight volunteers (students and junior researchers at IHMC) were asked to use the Explanation Satisfaction Scale to rate "good" and "bad" explanations. We calculated a 2 (Explanation Quality: "Good" vs. "Bad") x 2 (Object to be explained) mixed model ANOVA, for which the scale ratings were treated as a repeated measure. The main effect for ratings of Explanation Quality was only marginally significant [$F (1,6) = .455; p < .08$], but the size of the effect was quite large (Cohen's $d = 1.5$, equivalent to a $r^2 = .36$) indicating that, in general, scale ratings corresponded to higher values being awarded to good explanations ($M = 42.75; SD = 3.40$) and lower values being given for bad explanations ($M = 30.00; SD = 11.46$). No statistically significant effect was found for giving repeated scale ratings or for the interaction with Explanation Quality.

In conclusion, results from the content validity analysis and discriminant validity analysis show that the ES scale is valid. The Explanation Satisfaction Scale is presented in Appendix C. Like the Explanation Goodness Checklist (Appendix A), the Explanation Satisfaction Scale was based on the literatures in cognitive psychology, philosophy of science, and other pertinent disciplines regarding the features that make explanations good. Thus, the Explanation Satisfaction Scale is very similar to the Explanation Goodness Checklist. However, the application context for the two scales is quite different.

- The Explanation Goodness Checklist is intended to be used by researchers or their XAI systems that have created explanations. The Explanation Goodness Checklist is intended for use as an independent evaluation of explanations by other researchers. The reference is to the properties of explanations.
- Explanation Satisfaction is an evaluation of explanations by users. The Explanation Satisfaction Scale is for collecting judgments by research participants after they have worked with the XAI system that is being explained, and have been the beneficiaries of one or more explanations.

# 3. Measuring Mental Models

A methodology is needed for eliciting, representing, and analyzing users' mental models of intelligent systems or decision aid systems. In cognitive psychology, "mental models" are representations or expressions of how a person understands some sort of event, process, or system (Klein and Hoffman, 2008). In the XAI context, a mental model is basically a user's understanding of the AI system.

## 3.1 Users' Models of the Computer ("Mental Models") versus the Computer's Models of Users ("User Models")

In current parlance, the user's "mental model" is distinguished from what is called a "user model." The latter is a computer model of a user's mental model. The AI employs user models to adapt its operations and interactions. User models might consist of formalisms, such as production rules and semantic networks, which may represent a person's preferences and emulate some aspects of his or her knowledge and reasoning about some situation. User models for intelligent tutoring systems were aimed at understanding a learner's understanding so as to tailor instructional materials and exercises (see for instance, Lesgold, et al., 1992). User models are also referenced in the design of human-machine systems (see for instance, Carberry, 1990; Harris and Helander, 1984; May, Barnard, and Blandford, 1993). User models for AI and expert systems were aimed at emulating human cognition and thereby replicating performance on various tasks (see for instance, Anderson et al., 1990; Clancey, 1984, 1986). Mental models, including the person's beliefs and conceptual processes, may be represented as computational cognitive models using similar formalisms.

In this report, we do not review attempts to automatically generate user models (that is, formal instantiations of mental models). Friedman, Forbus and Sherin (2017) present present a computational cognitive model that creates and evaluates models based on fragmentary and inconsistent information. Their formalism is predicate calculus, and the assertion is that the system is capable of abductive inference, that is, inference to a best explanation. Our focus is on methods for eliciting information about users' mental models.

## 3.2 Empirical Assertions

There is a large body of research by experimental psychologists on mental models (e.g., Carroll, 1984; Gentner and Gentner, 1983; Greeno, 1977; Johnson-Laird, 1980, 1989; Kintsch, Miller & Poulson, 1974). Praetorious and Duncan (1988) present an elegant treatment of methodology for eliciting mental models. Staggers and Norco (1992) provide a good summary of the conceptual and theoretical issues.

There is a consensus that mental models can be inferred from empirical evidence. People may not be able to tell you "everything" about their understanding, and they may not be able to tell it well. But with adequate scaffolding by some method of guided task reflection, people can tell you how they understand an event or system, they can describe their knowledge of it, and the concepts and principles that are involved. There is sufficient evidence to conclude that different methods for eliciting mental models can converge (Evans, et al., 2001; van der Veer, & Melguzio, 2003).

For many people, mental imagery is involved in reasoning and dynamical and complex systems (e.g., Bogacz & Trafton, 2004). This assertion is supported by studies showing that diagrams can contribute significantly to the understanding of dynamic and complex systems (e.g., Clement, 2004; Glenberg and Langston, 1992; Heiser & Tversky, 2004; Johnson-Laird, 1983; Klein & Hoffman, 2008; Qin & Simon, 1992; Zhang & Wickens, 1987).

These ideas pertain directly to how people understand machines, spanning simple devices, process control systems, complex computational systems, and intelligent systems (Bainbridge, 1979, 1988; de Kleer and Brown, 1983; Goodstein, Andersen and Olsen, 1988; Moray, 1987; Mueller & Klein, 2011; Rasmussen, 1986; Rasmussen, Pejtersen, & Goodstein, 1994; Samurcay and Hoc, 1996; Staggers and Norcio, 1993; Williams, Hollan, and Stevens, 1983; Young, 1983).

## 3.3 Overview of Mental Model Elicitation Methods

Referring again to the historical differences between concepts of user models, cognitive models, and mental models, there are associated differences in how models of human cognition are elicited and represented. There are also differences in the emphasis placed on elicitation methods. For example, some of the work on user and cognitive models focuses on formal methods and languages for modeling beliefs and/or problem solving (e.g., procedural rules), These are researchers' computational models of what a user's mental model might be (see for instance, Clancey, 1986; Johnson-Laird, 1983; de Kleer, Doyle, Steele, and Sussman, 1977; Doyle, et al., 2002; Moray, 1983; Schaffernicht and Groesser, 2011).

Table 4 lists some methods that have been used to elicit mental models in fields such as cognitive psychology, knowledge acquisition for expert systems, instructional design, and educational computing. Table 5 lists some strengths and weaknesses of various methods.

Table 4. Some methods that can be used to elicit mental models.

| METHOD | ILLUSTRATIVE REFERENCES |
|---|---|
| Think-Aloud Problem Solving Task, in which participants think aloud during a task. | Reports by Pjtersen and Goodstein (1994) and Williams, Hollan and Stevens, (1983) are good illustrations of the task to elicit mental models specifically of devices. See also Beach, 1992; Ericsson and Simon, 1984; Gentner and Stevens, 1983; Greeno, 1983; Rasmussen, Pejtersen & Goodstein, 1994; Ward, et al., in press. |
| Think-Aloud Task with Concurrent Question Answering. | Gentner and Gentner, 1983; Williams, Hollan and Stevens, 1983 |
| Task Reflection or Retrospection Task, in which participants describe their reasoning after conducting a task (e.g., fault diagnosis). The retrospection can be based, for example, on a replay of their task performance such as in a video. | Fryer (1939) provides a clear and succinct presentation of a method that combines retrospection with Likert scale questionnaire to quantify participants' reasoning. See also Frederick, 2005; Lippa, Klein & Shalin, 2008; Praetorious & Duncan, 1988 |
| Structured interview, essentially Retrospection Task with Question-Answering. | Friedman, Forbus & Sherin, 2017; Fryer, 1939 |
| Card Sorting Task (also "Pathfinder"), based on the semantic similarity among a set of domain concepts. | The review article by van der Veer & Melguzio (2003) highlights this method. See also Chi, Feltovich & Glaser, 1981; St.-Cyr and Burns, 2002; Evans, et al., 2001; van der Veer & Melguzio, 2003 |
| Nearest Neighbor Task, in which participants select the explanation or diagram that best fits their beliefs. | Hardiman, Dufresne and Mestre, 1989; Klein and Militello, 2001 |
| Self-Explanation Task (also called "Teach-back"), in which the user/learner expresses their own understanding. Similar to the Retrospection/Reflection Task | Cañas et al., 2003; Ford, Cañas, & Coffey, 1993; Fermbach, et al., 2010; Molinaro & Garcia-Madruga, 2011; van der Veer & Melguzio, 2003 |
| Glitch Detector Task (also called "accident-error analysis"), in which people identify the things that are wrong in an explanation. | Hoffman, Coffey, Ford & Carnot, 2001; Taylor, 1988 |
| Prediction Task, in which users are presented test cases and are asked users to predict the results and then explain why they thought the predicted results would obtain. | Muramatsu & Pratt, 2001 |
| Diagramming Task (also "Concept Mapping"), in which users create a diagram that lays out their knowledge of processes, events and other relations. | Cañas, et al., 2003; Evans, et al., 2001; Hoffman et al., 2017; Moon, et al., 2011; Novak & Gowin, 1984 |
| ShadowBox Task, in which learners compare their understandings to those of a domain expert. | Klein & Borders, 2016 |

Table 5. Methods strengths and weaknesses.

| METHOD | STRENGTH | WEAKNESS |
|---|---|---|
| Concurrent Think-Aloud Problem Solving | Can provide rich information about mental models. | Transcription and protocol analysis can be time consuming, result in a great deal of data, require much analysis. |
| Think-Aloud Task with Concurrent Question Answering | Enables the researcher to present targeted probes to the user during task performance. | Highly dependent on the researcher/interviewer's skill at question design and interviewing. |
| Task Reflection or Retrospection | Can be conducted as a structured interview, as a questionnaire task, or as a post-task verbalization of a self-explanation. Can provide rich information about mental models, or a quick window into mental models. The process of self-explanation itself has learning value. | People can overestimate how well they understand complex systems. Transcription and protocol analysis can be time consuming, result in a great deal of data, require much analysis. Less effort would be required in a questionnaire method, though questionnaire design would be non-trivial. |
| Card Sorting | Can provide information about domain concepts and their relations. | Can provide sparse data about events or processes. Primary data consists only of similarity ratings (i.e., semantic nets) |
| Nearest Neighbor | Can provide a quick window into mental models. | People overestimate how well they understand complex systems. |
| Glitch Detector | Can support users to discover and explain aspects of their mental model that are reductive or incorrect. | Glitches have to be built-in. Knowledge shields may inhibit the awareness of reductive tendencies. |
| Question-Answering/Structured Interview | Enables researcher to probe selected aspects of a user's mental model. | Highly dependent on the researcher/interviewer's skill at question design and interviewing. |
| Prediction Task | Can provide a quick window into mental models. The predictions should be accompanied by a confidence rating and an free response elaboration that explains or justifies the predictions. | The free responses require content analysis. Requires clear rationale for the choice of instances or cases to be the focus of the predictions. |
| Diagramming Task | Can provide a rich and thorough representation of the user's mental model. Relations are not restricted to similarity (see Card Sorting Task, above). | Can take time to create, although user friendly software systems are readily available. |
| Box or "Shadowbox Lite" Task | Can provide a quick window into mental models. | May not result in a thorough expression of the mental model. |

## 3.4 Application to the XAI Context

What is most desirable for XAI is a task that enables the researcher to learn what is good and what is not good about a user's mental model, and enable the learner to learn what is good and what is limited in their understanding. An insight can be thought of as the self-awareness of a "knowledge shield" (Feltovich, Coulson, and Spiro, 2001). Knowledge shields are arguments that learners make that enable them to preserve their reductive understandings. A focus for instructional design has been to develop methods to get people to recognize when they are employing a knowledge shield that prevents them from developing richer mental models (Hilton, 1986, 1996; Prietula, Feltovich, & Marchak, 2000; Schaffernicht & Groesser, 2011; Tullio, Dey, Chalecki, & Fogarty, 2007).

What is most desirable for XAI is a method that can elicit mental models quickly, and result in data that can be easily scored, categorized, or analyzed.

It is important that the method provide some sort of structure or "scaffolding" that supports the user in explaining their thoughts and reasoning. One method is Cued Retrospection. Probe questions are presented to participants about their reasoning just after the reasoning task has been performed (see Ward, et al, in press). For instance, they might be asked, "Can you describe the major components or steps in the [software system, algorithm]." The probes can also reference metacognitive processes, for example by asking, "Based on the explanation of the [software, algorithm], can you imagine circumstances or situations in which the [software, algorithm] might lead to error conditions, wrong answers, or bad decisions?"

Instances of explanation are often expressions of what will happen in the future (Koehler, 1991; Lombrozo and Carey, 2006; Mitchell, Russo & Pennington, 1989). Prediction Tasks have the user predict what the AI will do, such as the classification of images by a Deep Net. A prediction task might be about "What do you think will happen next? but it can also be about "Why wouldn't something else happen?"  Explanation and counterfactual reasoning are co-implicative. Instances of explanation often are expressions of counterfactual reasoning (Byrne, 2017). A prediction task might serve as a frugal method for peering into users' mental models, but its application should be accompanied by a confidence rating and a free response elaboration in which the users explain or justify their predictions, or respond to a probe about counterfactuals

Diagramming can be an effective way for a user to convey the understanding to the researcher, and it also supports an analytical process in which diagrams are analyzed in terms of their proposition content (Cañas, et al., 2003; Johnson-Laird, 1983 Ch. 2). A diagramming task, along with analysis of concepts and relations (including causal connections or state transitions) has been noted in education as a method for comparing the mental models of students to those of experts or their instructor (see Novak and Gowin, 1984); noted in social studies as a method for comparing individuals' mental models of social groups (i.e., individuals and their inter-relations; Carley and Palmquist, 1992), and in operations research as a method for comparing mental models of dynamical systems (see Schaffernicht and Groesser, 2011).

Explicit Explanation improves learning and understanding. This holds for both deliberate self-motivated self-explanation and also self-explanation that is prompted or encouraged by the instructor. Self-explanation has a significant and positive impact on understanding. Self-generated deductions and generalizations help learners refine their knowledge (Chi and VanLehn, 1991: Chi, et al, 1989, 1994; Chi & Van Lehn, 1991; Lombrozo, 2016; Molinaro & Garcia-Madruga, 2011; Rittle-Johnson, 2006). Having learners explain the answers of "experts" also enhances the learner's understanding (Calin-Jagerman & Ratner, 2005).

This said, people sometimes overestimate how well they understand complex causal systems. This can be corrected by asking the learner to explicitly express their understanding or reasoning (Bjork et al., 2013; Chi, et al, 1989; 1991; Fernbach, et al., 2012; Mills & Keil, 2004; O'Reilly, Symons, & MacLatchy-Gaudet, 1998; Rittle-Johnson, 2006; Rosenblit & Keil, 2002; VanLehn, Ball, & Kowalski, 1990). "ShadowBox Lite" (presented by permission from Devorah Klein; see also Klein & borders, 2016) is a specific self-explanation task that is applicable to the XAI context, and may provide a quick window into user mental models: It avoids the necessity of eliciting and analyzing an extensive recounting of the user's mental model. In the method, the user is presented a question such as "How does a car's cruise control work?" Accompanying the question is a proposed explanation. The task for the user is to identify one or more ways in which the explanation is good, and ways in which it is bad. After doing this, the participant is shown a Good-Bad list that was created a domain expert. The participant's comparison of the lists can lead to insights.

### 3.5 Design Considerations

Referencing these four kinds of tasks, and the other tasks listed in Table 1, it is recommended that the evaluation of user mental models within the XAI context should employ more than one method for eliciting mental models. Performance on any one type of task might not align with performance at some other task. For instance, adequacy of a user-generated diagram did not match to better performance at a prediction task (see St.-Cyr and Burns, 2002). Performance on a simulated industrial process control task can be good and yet the operator's understanding can be limited and even incorrect (see Berry and Broadbent, 1984).

Not all of the participants in a study have to be presented with a mental model elicitation task. Indeed, a reasonably sized and representative set of 10 to 12 participants can be presented one or more mental model elicitation tasks. If the analysis of the goodness of the mental models aligns with measures of performance, then subsequent studies might use performance measures as a surrogate for mental model analysis.

### 3.6 Analysis of User Mental Models

An empirically-derived expression of the content or the ebbs and flows that compose a user's mental model must permit evaluation of mental model goodness (i.e., correctness, comprehensiveness, coherence, usefulness). For illustrative purposes, Table 3 presents a simple example of an Evaluation that is usually based on proposition analysis. Carley and Palmquist (1992) provide illustrative examples of propositional coding for transcripts of interviews of students by their teachers. The user model developed by Friedman, Forbus and Sherin (2017) is based on propositional encoding of interview transcripts. In this report we illustrate propositional analysis using the framework of the ShadowBox Lite Task. Table 6 presents an "expert" explanation, although this might just as easily be thought of as an explanation generated by an XAI system.

Table 6.  An example of propositional coding using the ShadowBox Lite Task.

| The control unit detects the rotation of the drive shaft from a magnet mounted on the drive shaft, and from that can calculate how fast the car is going. The control unit controls an electric motor that is connected to the accelerator linkage. The cruise control adjusts the engine speed until it is disengaged. ||
|---|---|
| What is right and helpful about this explanation? | What is problematic or wrong about this explanation? |
| The cruise control unit has to know how fast the car is going. | It seems overly technical, with some concepts left unexplained. |
| The cruise control has to control the engine throttle or accelerator. | I do not think the cruise control detects the engine speed. |

Results from a number of elicitation tasks will generally be sets of sentences, which can be recast as propositions, broken out by the concepts — the relations. The products from a diagramming task can also be recast as a set of propositions. The explanation can be decomposed into the component concepts, relations and propositions. In the case of the cruise control example, the expert's explanation has ten concepts (drive shaft rotation, car speed, etc.) seven relations (mounted on, disengage, etc.), and six propositions (e.g., Magnet is mounted on the drive shaft, Control unit calculates car speed).

The counts for concepts, relations and propositions can be aggregated and analyzed in a number of ways. For instance, one can calculate the percentage of concepts, relations, and propositions that are in the user's explanation that are also in the expert's explanation. This could suggest the completeness of the user's mental model.

## 3.7 Hypothesis Testing

Based on the model in Figure 1, a primary hypothesis for XAI is that a measure of performance is simultaneously a measure of the goodness of user mental models. As our discussion (above) of the multi-method approach suggests, this hypothesis is not to be taken for granted. It can be empirically investigated to see if it is confirmed. It may be especially revealing to compare results from participants who perform the best and participants who perform the worst. Comparison of their models, and those with an expert model would affirm the hypothesis that a measure of performance can be simultaneously a measure of the goodness of the user mental model.

# 4. Measuring Curiosity

## 4.1 Introduction

There are theoretical and empirical reasons why curiosity could be considered an important factor in Explainable AI. One of the most important reasons is that the act of seeking an explanation is driven by curiosity. Therefore, it is important that the XAI systems harness the power of curiosity. Explanations may promote curiosity and set the stage for the achievement of insights and the development of better mental models.

On the other hand, explanations can actually suppress curiosity and reinforce flawed mental models. This can happen in a number of ways:

- An explanation might overwhelm people with details,
- The XAI system might not allow questions or might make it difficult for the user to pose questions,
- Explanations might make people feel reticent because of their lack of knowledge,
- Explanations may include too many open variables and loose ends, and curiosity decreases when confusion and complexity increase.

For these reasons, the assessment of users' feelings of curiosity might be informative in the evaluation of XAI systems.

## 4.2 The Nature of Curiosity

"Epistemic curiosity" is the general desire for knowledge, a motive to learn new ideas, resolve knowledge gaps, and solve problems, even though this may entail effortful cognitive activity (Berlyne, 1960, 1978; Loewenstein, 1994; see also Litman, 2008; 2010). Stimulus novelty, surprisingness, or incongruity, can trigger curiosity. All of these features refer to circumstances when information is noted as being missing or incomplete.

Curiosity is also triggered in circumstances where one experiences a "violated expectation" (Maheswaran & Chaiken, 1991). Violated expectations essentially reflect the discovery that events anticipated to be comprehensible are instead confusing – more information and some sort of change to one's understanding will be needed to make sense of the event and thus resolve the disparity between expectation and outcome. Such situations lead people to engage in effortful processing, and motivates them to seek out additional knowledge in order to gain an insight and resolve the incongruency (Loewenstein, 1994).

This is directly pertinent to XAI. Curiosity is stimulated when learner recognizes that there is a gap in their knowledge or understanding. Recognizing a knowledge gap, closing that gap, and achieving satisfaction from insight make the likelihood of success from explanations or self-explanations seem feasible. This leads to the question of how to assess or measure curiosity in the XAI context, and what to do with the measurements.

Figure 2 presents a conceptual model of the explanation process from the perspective of the learner who is using an XAI system. This diagram shows the role and place of curiosity, noting that some learners may not feel curious.

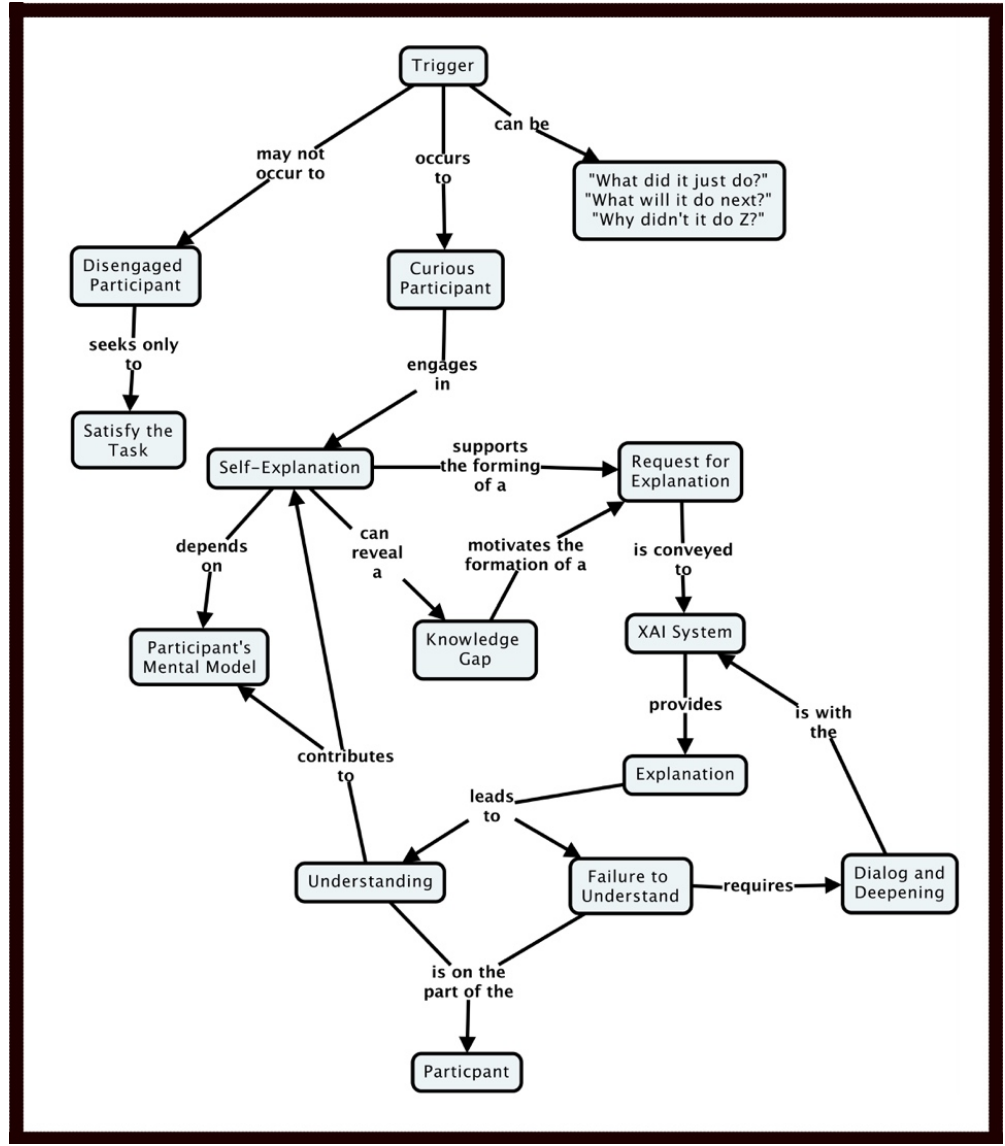

Figure 1. A model of the explanation process in the XAI context from the perspective of the learner, showing the role of curiosity.

## 4.3 Measuring Curiosity in the XAI Context

A number of psychometric instruments have names that at first glance make them pertinent to XAI, such as Frederick's "Cognitive Reflection Test" (2005). But this taps numerical fluency and competence. Available scales of curiosity, such as Kashdan's "Curiosity Exploration Inventory"(Kashdan, et al., 2004), the Cacioppo "Need for Cognition" scale (Cacioppo, Petty & Kao, 1984), and Litman's "I-Type/D-Type Curiosity Scales" (Litman and Jimeson, 2004) consider curiosity to be a pervasive style or personality trait. As such, the instruments ask questions such as:
> I actively seek as much information as I can in a new situation,
> I feel stressed or worried about something I do not know,
> I like to discover new places to go.

Such items are barely applicable in the XAI context, in which curiosity is situation or task specific, and refers to the workings of computational devices, rather than daily life or experiences.

The above discussion of the knowledge gap theory of curiosity implies that the evaluation of XAI systems can benefit from basically asking users to identify the triggers that motivated them to ask for an explanation. As for the other measurement classes we have discussed, it is valuable for the curiosity measurement method to present a "quick window." Table 7 presents a simple questionnaire that can be administered to research participants whenever they ask for an explanation.

Table 7. A Curiosity Checklist.

| Why have you asked for an explanation? Check all that apply. |
|---|
| I want to know what the AI just did. |
| I want to know that I understand this AI system correctly. |
| I want to understand what the AI will do next. |
| I want to know why the AI did not make some other decision. |
| I want to know what the AI would have done if something had been different. |
| I was surprised by the AI's actions and want to know what I missed. |

Responses will be informative with regard to these things:
(1). Responses may serve as parameters or constraints that the XAI system uses to generate explanations.
(2) Responses may provide a window into aspects of the AI system's operations that need explaining.
(3). Responses may reveal ways in which the AI's explanation method might suppress or inhibit curiosity, and
(4). Responses may make it possible to use depth of curiosity (i.e., more triggers are checked) as an independent variable.

# 5 Measuring Trust in the XAI Context
## 5.1 Introduction

Trust in automation is of concern in computer science and cognitive systems engineering, as well as the popular media (e.g., Chancey et al., 2015; Hoff and Bashir 2015; Fitzhugh, Hoffman & Miller, 2011; Hoffman et al., 2009, 2013; Huynh et al., 2006; Kucala, 2013; Lee & See, 2004; Merritt and Ilgen, 2008; Merritt et al. 2013, 2015; Naone, 2009; Pop et al. 2015; Shadbolt, 2002; Wickens et al., 2015; Woods and Hollnagel 2006).

Some users may take the computer's assertions (data, claims) as valid and true because they come from a computer. But other users may require some sort of justification – empirical reasons to believe that the computer's presentations or assertions are valid and true. Just as there are varieties of trusting, there are varieties of negative trusting (mistrust, distrust, etc.). Trust in automation can rapidly break down under conditions of time pressure, or when there are conspicuous system faults or errors, or when there is a high false alarm rate (Dzindolet et al., 2003; Madhavan and Wiegmann 2007). The trusting of machines can be hard to reestablish once lost.

However trust is measured, in the XAI context, the measurement method must be sensitive to the emergence of negative trusting states. XAI systems should enable the user to know whether, when, and why to trust and rely upon the XAI system and know whether, when, or why to mistrust the XAI and either not rely upon it, or rely on it with caution. People always have some mixture of justified and unjustified trust, and justified and unjustified mistrust in computer systems. A user might feel positive trust toward an AI system with respect to certain tasks and goals and simultaneously feel mistrusting or distrusting when other tasks, goals or situations are involved. Indeed, in complex sociotechnical work systems, this is undoubtedly the norm in the human-machine relation (Hoffman et al. 2014; Sarter et al. 1997),

Ideally, with experience the user comes to trust the computer with respect to certain tasks or goals in certain contexts or problem situations and appropriately mistrusts the computer with respect to certain tasks or goals in certain contexts or problem situations. Only if this trusting relation is achieved can the user's reliance on the computer be confident (Riley, 1996).

## 5.2 Explanation as Exploration

Trusting of XAI systems will always be exploratory. Active exploration of trusting–relying relationships cannot and should not be aimed at achieving single stable states or maintaining some decontextualized metrical value, but must be aimed at maintaining an appropriate and context-dependent expectation. Active exploration will involve:
- Verifying the reasons to take the computer's presentations or assertions as true,
- Enabling the user to understand and anticipate circumstances in which the machine's recommendations will not be trustworthy, and the computer's recommendations should not be followed even though they appear trustworthy.

- Assessing the situational uncertainty that might affect the probability of favorable outcomes.
- Enabling the user to identify and mitigate Unjustified Trusting and Unjustified Mistrusting situations.
- Enabling the user to discover and appreciate indicators to mitigate the impacts and risks of unwarranted reliance, or unwarranted rejection of recommendations, especially in time-pressured or information-challenged (too much, too little, or uncertain) situations.
- Enabling the user to understand and anticipate circumstances (i.e., unforeseen variations on contextual parameters) in which the XAI should not be trusted even if it is working as it should, and perhaps especially if it is working as it should (Woods 2011).

There are many possible trajectories in this exploration. For example, a user who is initially cautions or skeptical may benefit from a good explanation and move into a region of justified trust. Subsequent use of the XAI system, however, may result in an automation surprise. For example, a Deep Net might make a misclassification that no human would make. This might swiftly move the user into a state of unjustified mistrust, in which the user is skeptical of any of the classifications that the Deep Net makes. Following that, the XAI system may provide additional explanations and the user may explore the performance of the XAI, converging in the region of appropriate trust and reliance.

## 5.3 Trust Measurement Scales

The scientific literature on trust (generally) presents a number of scales for measuring trust. The majority of trust scales have been developed for the context of interpersonal trust. We focus on scales designed for use in the assessment of human trust in automation. A synopsis of representative trust scales is presented in Appendix D.

Minimally, a trust scale can ask two questions: Do you trust the machine's outputs? (trust) and Would you follow the machine's advice? (reliance). Indeed, these two items comprise the scale developed by Adams, et al. (2003). The scale developed by Johnson (2007) asks only about reliance and the rareness of errors.

Some scales for assessing trust in automation are highly specific to particular application contexts. For example, the scale developed by Schaefer (2013) refers specifically to the context of human reliance on a robot, and thus asks Does it act as part of a team? and Is it friendly? As another example, the trust in automation scale developed by Adams, et al. (2003) refers specifically to the evaluation of simulations. It asks only two questions, one about trust (Do you trust it?) and one about reliance (Are you prepared to rely on it?), although it is noteworthy that this is the only scale that has a free response option associated with the two scale item questions.

To be sure, some research on trust in automation is clearly inapplicable to the XAI context. For example, Heerink, et al. (2010) were interested in the acceptance of an assistive/robotic technology by the elderly. The questionnaire they utilized has such items as I feel the robot is nice, and The robot seems to have real feelings.

Montague (2010) presented a study aimed at validating a scale for trust in medical diagnostic instruments, but all of the items actually refer to trust in the health care provider and positive affect about the provider. Abstracting from that reference context, the other items ask about reliability, correctness, precision, and trust. Thus, we see essential similarity to the items in the Cahour-Fourzy Scale.

Some scales for assessing trust in automation are highly specific to particular experimental contexts. Hence, the items are not applicable to the XAI context, or to any generic trust-in-automation context. For example, the scale by Dzindolet, et al. (2003) was created for application in the study of trust in a system for evaluating terrain in aerial photographs, showing images in which there might be camouflaged soldiers. Thus, the hypothetical technology was referred to as a "contrast detector." The experiment was one in which the error rate of the hypothetical detector was a primary independent variable. As a consequence, the scale items refer to trials e.g., How well do you think you will perform during the 200 trials? (Not very well-Very well), and How many errors do you think you will make during the 200 trials? Some of the scale items can be adapted to make them appropriate to the XAI context, but the result of this modification is just a few items, which are ones that are already in the Cahour-Fourzy Scale (e.g., Can you trust the decisions the [system] will make?)

Of those scales that have been subject to psychometric analysis of reliability, results suggest that trust in automation scales can be reliable. Of those scales that have been subject to validity analysis, high Cronbach alpha results have been obtained. The report by Jian, et al. (2000) illustrates these psychometric analyses.

## 5.4 Recommendations

Looking across the various Scales (see Appendix D), there is considerable overlap, and cross-use of the scale items. We have distilled a set of items that might be used in XAI research. This is presented in Appendix E. Most of the items are from the Cahour-Fourzy scale (some of which are also in the Jian et al. scale), but the recommended scale incorporates items from other scales.

None of the scales that have been reviewed treat trust as a process; they treat it as a static quality or target state, that is measured after the research participants have completed their experimental tasks. In contrast, it is recommended for the XAI Program that trust measurement be a repeat measure. The scale or selected scale items can be applied after individual trials (e.g., after individual XAI categorizations or recommendations; after individual explanations are provided, etc.). The full scale could be completed part way through a series of experimental trials, and at the conclusion of the final experimental trial. Multiple measures taken over time could be integrated for overall evaluations of human–machine performance, but episodic measures would be valuable in tracking such things as: How do users maintain trust? What is the trend for desirable movement toward appropriate trust?

# 6 Measuring Performance

The goal of performance measurement is to determine the degree of success of the human-machine system at effectively conducting the tasks for which the technology is designed. Based on the conceptual model in Figure 1, the key hypotheses are:

- User performance (including measures of joint user-system performance) will improve as a result of being given satisfying explanations.
- User performance will be a function of the qualities of their mental model (e.g., correctness, completeness, etc.)
- User performance may be affected by their level of epistemic trust
- User performance (that is, reliance) will be appropriate if the user has been able to explore the competence envelop of the AI system.

The evaluation of the performance of an XAI system cannot be neatly divorced from the evaluation of the performance of the user, or from the performance of the human-machine system as a whole.

## 6.1 Performance with Regard to the Primary Task Goals

The AI system will have some primary task goal. This might be to correctly categorize images, correctly identify certain kinds of actions in videos, or conduct an emergency rescue operation. Performance can be measured in terms of the number of trials on which the user met with success, within some pre-specified time period. In a search-and-rescue use case, work system performance might be measured in terms of the number of trials that a user has to work in order to reach some pre-determined criterion. Basic measures of efficiency can be applied, expressing the ratio of the number of tasks or sub-tasks completed per some period of time.

## 6.2 Performance with Regard to the User

A second aspect of performance measurement is the quality of the performance of the user, such as the correctness of the user's predictions of what the AI will do. For these aspects of performance, just like performance with regard to primary ask goals, one can measure response speed and correctness (hits, errors, misses, false alarms), but in this case the user's predictions of the machine outputs. Examination can be made for both typical and atypical cases/situations. Additionally, one can measure the correctness and completeness of the user's explanation of the machine's output for cases that are rare, unusual, or anomalous.

## 6.3  Performance with Regard to the Work System

With regard to performance at the work system level, considerations other than raw efficiency come into play (Hoffman & Hancock, 2017; Koopman & Hoffman, 2003). For example, an XAI system that drives the work efficiency (say, to deal with data overload issues) may not make for a very contented workplace (Merritt, 2011). The complexity of performance at the work system level requires analysis based on trade-off functions (Woods & Hollnagel, 2006).

Work system level performance would be reflected when scores on a measure of explanation satisfaction correlate highly with evaluations of the goodness of the users' mental models or with scores on some other performance measure.

At the work system level, appropriate trust and reliance emerge from the user's experience, especially as the user encounters tough cases or cases that fall at the boundary of the work system's competence envelope. Appropriateness refers to the fact that mistrust, as well as trust, can be justified. Presumably, appropriate reliance (knowing when and when not to rely on the system's outputs) hinges on appropriate trust.

The analysis of work system level performance may employ some measure of controlability, that is, the extent to which the human can produce an intended set of outcomes based on given inputs or conditions. Analysis may employ some measure of correctability. This is a measure of the extent or ease with which the user can correct the machine's activities so as to make the machine outputs better aligned with either objective states of affairs or the user's judgments of what the machine should be determining.

The analysis of work system level performance can involve comparing the work productivity of the XAI-User work system to productivity in current practice (baseline). A related method is to look at learning curves. This involves establishing a metric on a productivity scale, a metric that identifies when performance is satisfactory. How many trials of test cases must a user work successfully in order to reach that learning criterion? Is the learning rapid?  Why do some people take a long time to reach criterion? The advantage of a trials-to-criterion approach to measurement is that it could put the different XAI systems on a "level playing field" by making the primary measure a derivative of task completion time. A variation on this method is to compare performance in this way with performance when the Explanation capability of the XAI is somehow hobbled.

Perhaps the most powerful and direct way of evaluating the performance of a work system that includes an XAI is to evaluate how easy or difficult it is to get prospective users (stakeholders) to adopt the XAI system. In discussing early medical diagnosis systems, van Lent et al (2004) stated, "Early on the developers of these systems realized that doctors weren't willing to accept the expert system's diagnosis on faith" (p. 904), which led to development of the first explanation-based AI systems.  Many users may be satisfied with shallow explanations in that they will be willing to adopt the system that has "XAI" or may prefer it over another non-XAI system when making adoption decisions. Measures of choice behavior were also advocated by Adams et al. (2003) as a way of measuring human trust in automation.

# 7. Prospects

Measurement is the foundation of empirical inquiry. As a form of exploration, one of the purposes of making measurements is to improve the measures. We do not regard the ideas and methods discussed in this paper as being final.

It was from our own empirical inquiry that emerged our distinction between explanation goodness as an *a priori* evaluation by researchers versus explanation satisfaction as an *a posteriori* evaluation by learners.

We look forward to refinements and extensions of all the ideas presented in this article.

Also foundational to empirical inquiry is a theoretical foundation. The one on which we have relied is the conceptual model presented in Figure 1 (above), which references measures of explanations, user models, and performance. Other measurable theoretical concepts might be of value for XAI research and development. As our discussion suggests, we advocate for a multi-method approach. Such an approach seems mandated by the fact that the human-XAI interaction is a complex cognitive system.

An important measurement topic that we have not addressed is that of "metrics." In modern discourse, the word "metric" often carries a dual meaning. One is "measure" or "measurement." The other is the notion of some sort of threshold on a measurement scale that is used to make evaluations or decisions. In this article we have discussed measurement, not metrics. A simple example of the measure-metric distinction might clarify this. If a surgeon specializing in carpal tunnel syndrome has a success rate less than 90 percent, he or she might well be in trouble. On the other hand, a specialist in spinal surgery with a 90% success rate would be considered a miracle worker. The measurement scale in both contexts is surgery success rate. The metric that is laid on that scale depends on the application context (for a fuller discussion, see Hoffman, 2010).

Certainly XAI research needs metrics to accompany its measures. Is performance superior, acceptable, or poor? Is a mental model rich or impoverished? But metrics for these sorts of decisions do not emerge directly or easily from theoretical concepts being measured, or the operationalzed measures that are being used. The operational definition of a measure tells you how to make measurements, not how to interpret them. Metrics come from policy.

Resolution of this metrics challenge may emerge as more XAI projects are carried through all the way to performance evaluation.

XAI Metrics                                                                                           p. 25# 8. References

Adams, B.D., Bruyn, L.E., & Houde, S. (2003). Trust in automated systems. Report, Ministry of National Defence, United Kingdom.

Alang, N. (2017, August 31). Turns out algorithms are racist. *The New Republic*. Retrieved from https://newrepublic.com/article/144644/turns-algorithms-racist?utm_content=buffer7f3ea

Anderson, J., Boyle, C.F., Corbett, A.T., and Lewis, M.W. (1990). Cognitive modeling and intelligent tutoring. In W. J. Clancey and E. Soloway (Eds.) *Artificial intelligence and learning environments* (pp. 7–49). Cambridge, MA: Bradford Books.

Ayre, C., & Scally, A.J. (2014). Critical values for Lawshe's Content Validity Ratio: Revisiting the original methods of calculation. *Measurement and Evaluation in Counseling and Development, 47*, 79-86.

Bainbridge, L. (1979). Verbal reports as evidence of process operator's knowledge. *International Journal of Man-Machine Studies, 11*, 411-436.

Bainbridge, L. (1988). Types of representation. In I.P. Goodstein, H.B. Andersen, and S.E. Olsen, (Eds.) *Tasks, errors and mental models* (pp. 70-91). New York: Taylor and Francis.

Beach, L. R. (1992). Epistemic strategies on causal thinking in expert and nonexpert judgment. In G. Wright & F. Bolger (Eds.), *Expertise and decision support* (pp. 107-127). New York: Plenum.

Berlyne, D.E. (1960). *Conflict, arousal, and curiosity*. New York: McGraw-Hill.

Berlyne, D.E. (1978). Curiosity and learning. *Motivation and Emotion, 2*, 97-175.

Berry, D. C., and Broadbent, D. E. (1988). Interactive tasks and the implicit-explicit distinction. *British Journal of Psychology, 79*, 251–272.

Biran, O., & Cotton, C. (2017). Explanation and Justification in Machine Learning: A Survey. *IJCAI-17 Workshop on Explainable Artificial Intelligence (XAI)*. http://home.earthlink.net/~dwaha/research/meetings/ijcai17xai/1.%20(Biran%20&%20Cotton%20XAI-17)%20Explanation%20and%20Justification%20in%20ML%20-%20A%20Survey.pdf

Bjork, R.A., Dunlosky, J., & Kornell, N. (2013). Self-regulated learning: Beliefs, techniques, and illusions. *Annual Review of Psychology, 64*, 417–444.

Bogacz, S. and Trafton, J.G. (2004). Understanding dynamic and static displays: Using images to reason dynamically. *Cognitive Systems Research, 6*, 312-319.

Bornstein, A.M. (2016, September 1). Is Artificial Intelligence Permanently Inscrutable? Retrieved August 29, 2017, from http://nautil.us/issue/40/learning/is-artificial-intelligence-permanently-inscrutable

Byrne, R.M. (2017). Counterfactual Thinking: From Logic to Morality. *Current Directions in Psychological Science, 26*(4), 314–322.

Cacioppo, J. T., Petty, R. E., & Kao, C. F. (1984). The efficient assessment of need for cognition. *Journal of Personality Assessment, 48*(3), 306-307.

XAI Metrics p. 31Maheswaran, D., and Chaiken, S. (1991). Promoting systematic processing in low-motivation settings: Effect of incongruent information on processing and judgment. *Journal of Personality and Social Psychology, 61*,13-25.

May, J., Barnard, P.J., and Blandford, A. (1993). Using structural descriptions of interfaces to automate the modeling of user cognition. *User Modeling and User-Adapted Interaction, 3*, 27-64.

McKeown, K.R., & Swartout, W. R. (1987). Language generation and explanation. *Annual Review of Computer Science*, *2*(1), 401–449.

Merritt, S. M. (2011). Affective processes in human–automation interactions. *Human Factors:, 53*(4), 356-370.

Merritt, S.M., Heimbaugh, H., LaChapell, J., and Lee, D. (2013). I trust it, but don't know why: Effects of implicit attitudes toward automation in trust in an automated system. *Human Factors, 55*, 520–534.

Merritt, S.M., and Ilgen, D.R. (2008). Not all trust is created equal: Dispositional and history-based trust in human–automation interactions. *Human Factors, 50*, 194–201.

Merritt, S.M., Lee, D., Unnerstall, J.L., and Huber, K. (2015). Are well-calibrated users effective users? Associations between calibration of trust and performance on an automation- aided task. *Human Factors, 57*, 34–47.

Miller, T. (2017). Explanation in Artificial Intelligence: Insights from the Social Sciences. ArXiv:1706.07269 [Cs]. Retrieved from http://arxiv.org/abs/1706.07269

Mills, C. M., & Keil, F. C. (2004). Knowing the limits of one's understanding: The development of an awareness of an illusion of explanatory depth. *Journal of Experimental Child Psychology*, *87*(1), 1–32.

Mitchell, D., Russo, E., & Rennington, N. (1989). Back to the future: Temporal perspective in the explanation of events. *Journal of Behavioral Decision Making*, *2*, 25–38.

Molinaro, R.I., and Garcia-Madruga, J.A. (2011). Knowledge and question asking. *Psicothema, 23*, 26-30.

Montague, E. (2010). Validation of a trust in medical technology instrument. *Applied ergonomics, 41*(6), 812-821.

Moon, B.M., Hoffman, R.R., Cañas, A.J., and Novak, J.D. (Eds.) (2011). *Applied concept mapping: Capturing, Analyzing and Organizing Knowledge*. Boca Raton, FL: Taylor and Francis.

Moore, J. D., & Swartout, W. R. (1990). Pointing: A Way Toward Explanation Dialogue. In *Proceedings of AAAI*, *90*, 457–464. Menlo Park, CA: AAAI.

Moray, N. (1987). Intelligent aids, mental models, and the theory of machines. *International Journal of Man-Machine Studies, 7*, 619-629.

Mueller, S.T. and Klein, G. (2011, March/April). Improving users' mental models of intelligent software tools. *IEEE Intelligent Systems,* 77–83.

# Appendix A

# Explanation Goodness Checklist

This checklist is a list of the features that make explanations good, according to the research literature. *The reference is to the properties of explanations*.

The intended use context is for researchers or domain experts to provide an *independent, a priori* evaluation of the goodness of explanations that are generated by *other* researchers or by XAI systems.

---

The explanation helps me **understand** how the [software, algorithm, tool] works.

| YES | |
|---|---|
| NO | |

The explanation of how the [software, algorithm, tool] works is **satisfying**.

| YES | |
|---|---|
| NO | |

The explanation of the [software, algorithm, tool] sufficiently **detailed.**

| YES | |
|---|---|
| NO | |

The explanation of how the [software, algorithm, tool] works is sufficiently **complete.**

| YES | |
|---|---|
| NO | |

The explanation is **actionable**, that is, it helps me know how to use the [software, algorithm, tool]

| YES | |
|---|---|
| NO | |

The explanation lets me know how **accurate or reliable** the [software, algorithm] is.

| YES | |
|---|---|
| NO | |

The explanation lets me know how **trustworthy** the [software, algorithm, tool] is.

| YES | |
|---|---|
| NO | |

# Appendix B

## Materials used in the Evaluation of the Discriminant Validity
## of the Explanation Satisfaction Scale

---

### How do cell phones provide directions?

**Explanation Judged *a priori* to be Relatively "Good"**

Cell phones use the Global Positioning System of satellites to determine the exact location of the phone.

A web-based service (such as Google Maps, Apple Maps, etc.) identifies the phone's coordinates along with the desired destination entered by the phone user.

This information is transmitted to a computer, where The Map Service calculates the best route based on shortest path, but also traffic and other variables.

The instructions are then sent back to the phone.

The Map Service monitors the phone's location in real-time, and each step is given to the user based on the phone's proximity to the next step (or checkpoint).

---

**Explanation Judged *a priori* to be Relatively "Bad"**

Cell phones can provide good directions because they have really accurate maps on them.

The phone downloads the maps from the internet.

The phone knows where it is and it calculates the shortest route.

It tries to give step-by-step directions.

# How does automobile cruise control work?

**Explanation Judged *a priori* to be Relatively "Good"**

The speed of the car at the moment you turn on the cruise control is stored in a memory circuit.

The cruise control device reads the car's speed by a speed sensor that is on the car's drive shaft.

The cruise control is linked to the engine accelerator, and accelerates the car when it falls below the speed you set, such as when you start to go uphill, for example.

It can stop accelerating the car when the car is going downhill, but does not apply the brakes to decelerate the car.

The cruise control also senses when the brake pedal has been depressed, and disengages from the accelerator.

**Explanation Judged *a priori* to be Relatively "Bad"**

The vacuum control unit reads the pulse frequency from a magnet mounted on the drive shaft to figure out how fast the car is going.

A bidirectional screw-drive electric motor that is connected to the accelerator linkage receives the control signal and transmits it to the throttle.

When you set the cruise control, the engine detects the engine RPM at that point and tries to maintain that engine speed until it is disengaged.

The cruise control detects the car gear, which it adjusts to allow the engine to maintain a constant RPM.

| **How do computers predict hurricanes?** |
|---|
| **Explanation Judged *a priori* to be Relatively "Good"**<br><br>Computers have a mathematical model of the atmosphere that divides the world into many small regions, each just a few square kilometers.<br><br>Each region is defined in terms of its air pressure, temperature, winds, and moisture.<br><br>The computer calculates what will happen at the boundaries of each region. For example, strong winds in one region will move air into an adjacent region.<br><br>These calculations must be performed for every boundary between all the regions. This allows the prediction of the path a hurricane will take. |
| **Explanation Judged *a priori* to be Relatively "Bad"**<br><br>The computers have a database of all previous hurricanes and the paths that they followed.<br><br>Once a hurricane is located, using a satellite image, the computer accesses the database and determines the path that was most frequently taken by hurricanes having that initial location.<br><br>This process is repeated once every hour, tracking the hurricane as it moves.<br><br>The computers can also tell when the winds and rain will impact the land, and that is when the hurricane warnings are issued. |

# Appendix C

# Explanation Satisfaction Scale

1. From the explanation, I **understand** how the [software, algorithm, tool] works.

| 5 | 4 | 3 | 2 | 1 |
|---|---|---|---|---|
| I agree strongly | I agree somewhat | I'm neutral about it | I disagree somewhat | I disagree strongly |

2. This explanation of how the [software, algorithm, tool] works is **satisfying**.

| 5 | 4 | 3 | 2 | 1 |
|---|---|---|---|---|
| I agree strongly | I agree somewhat | I'm neutral about it | I disagree somewhat | I disagree strongly |

3. This explanation of how the [software, algorithm, tool] works has **sufficient detail.**

| 5 | 4 | 3 | 2 | 1 |
|---|---|---|---|---|
| I agree strongly | I agree somewhat | I'm neutral about it | I disagree somewhat | I disagree strongly |

4. This explanation of how the [software, algorithm, tool] works seems **complete**.

| 5 | 4 | 3 | 2 | 1 |
|---|---|---|---|---|
| I agree strongly | I agree somewhat | I'm neutral about it | I disagree somewhat | I disagree strongly |

5. This explanation of how the [software, algorithm, tool] works **tells me how to use** it.

| 5 | 4 | 3 | 2 | 1 |
|---|---|---|---|---|
| I agree strongly | I agree somewhat | I'm neutral about it | I disagree somewhat | I disagree strongly |

6. This explanation of how the [software, algorithm, tool] works is **useful to my goals**.

| 5 | 4 | 3 | 2 | 1 |
|---|---|---|---|---|
| I agree strongly | I agree somewhat | I'm neutral about it | I disagree somewhat | I disagree strongly |

7. This explanation of the [software, algorithm, tool] shows me how **accurate** the [software, algorithm, tool] is.

| 5 | 4 | 3 | 2 | 1 |
|---|---|---|---|---|
| I agree strongly | I agree somewhat | I'm neutral about it | I disagree somewhat | I disagree strongly |

8. This explanation lets me judge when I should **trust and not trust** the [software, algorithm, tool]

| 5 | 4 | 3 | 2 | 1 |
|---|---|---|---|---|
| I agree strongly | I agree somewhat | I'm neutral about it | I disagree somewhat | I disagree strongly |

# APPENDIX D

## Synopsis of Representative Trust Scales

**Cahour-Forzy (2009) Scale; Adams, et al. (2003) Scale**

Trust (and distrust) are defined as a sentiment resulting from knowledge, beliefs, emotions and other aspects of experience, generating positive or negative expectations concerning the reactions of a system and the interaction with it. The scale was developed in the context of learning to use a cruise control system. Trust was analyzed into three factors: reliability, predictability, and efficiency. The scale asks users directly whether they are confident in the XAI system, whether the XAI system is predictable, reliable, safe, and efficient.

The scale assumes that the participant has had considerable experience using the XAI system. Hence, these questions would be appropriate for scaling after a period of use, rather than immediately after an explanation has been given and prior to use experience. In the original scale, the items are rated on a bipolar scale going from "I agree completely" to "I do not agree at all." The items we present below have been slightly modified to fit the general Likert form developed for the XAI Explanation Satisfaction Scale. In addition to conforming to psychometric standards, consistency of format will presumably make the ratings tasks easier for participants.

1. What is your confidence in the [tool]? Do you have a feeling of trust in it?

| 1 | 2 | 3 | 4 | 5 | 6 | 7 |
|---|---|---|---|---|---|---|
| I do not trust it at all. | | | | | | I trust it completely |

2. Are the actions of the [tool] predictable?

| 1 | 2 | 3 | 4 | 5 | 6 | 7 |
|---|---|---|---|---|---|---|
| It is not at all predictable. | | | | | | It is completely predictable. |

3. Is the [tool] reliable? Do you think it is safe?

| 1 | 2 | 3 | 4 | 5 | 6 | 7 |
|---|---|---|---|---|---|---|
| It is not at | | | | | | It is |
| all safe. | | | | | | completely safe. |

4. Is the [tool] efficient at what it does?

| 1 | 2 | 3 | 4 | 5 | 6 | 7 |
|---|---|---|---|---|---|---|
| It is not at all efficient. | | | | | | It is completely efficient. |

Adams, et al. actually developed two scales. One was for the evaluation of simulations but the other was generic, intended for the evaluation of any form of automation. Apart from the item about liking, the items show overlap with items in the Cahour-Fourzy Scale.

Each item is accompanied by a bipolar rating scale (e.g., Useful-Not Useful; Reliable-Not reliable) on which the participant makes a tick mark on a -5 to +5 delineation. Following the Likert items, the Scale asks participants to rank the importance of the six item factors.

Is the automation tool useful?
How reliable is it?
How accurately does it work?
Can you understand how it works?
Do you like using it?
How easy is it to use?

**Jian, et al. Scale (2000)**

Trust is regarded as a trait. It is analyzed into six factors: Fidelity, loyalty, reliability, security, integrity, and familiarity. Factors were developed from cluster analysis on trust-related words. This scale is one of the most widely used, especially in the field of human factors. Indeed, a number of other scales have used items, or have adapted scale items, from the Jian, et al. Scale.

The item referencing "integrity" is problematic as the concept that a machine can act with integrity is not explicated. The final item, about familiarity, would not be relevant in the SAI context, since the participants' degree of experience with the XAI system will be known objectively.

Items 1, 2, 3, and 4 all seem to be asking the same thing.

The other items items in this scale show considerable overlap with items in the Cahour-Fourzy scale. However, item 4 is particularly interesting and is does not have a counterpart in the Cahour-Fourzy Scale. We are inclined to recommend that the Jian, et al., item 4 be incorporated into the XAI version of the Cahour-Fourzy Scale.

1. The system is deceptive.
2. The system behaves in an underhanded manner.
3. I am suspicious of the system's intent, action, or outputs.
4. I am wary of the system.
5. The system's actions will have a harmful or injurious outcome.
6. I am confident in the system.
7. The system provides security.
8. The system has integrity.
9. The system is dependable.
10. I can trust the system.
11. I am familiar with the system.

**Madsen-Gregor Scale (2000)**

Trust is defined as being both affective and cognitive. Trust was analyzed into five factors: reliability, technical competence, understandability, faith, and personal attachment. Their focus was not just trust in a decision aid but trust in an intelligent decision aid. As such, their scale deserves our particular attention. Unfortunately, reports on their work are not accompanied by information about the precise method for administering the scale (i.e., whether or not it used a Likert method). That said, their results show very high reliabilities (alpha = 0.94) and a factor analysis that accounts for about 70% of the variance.

| | |
|---|---|
| Perceived Reliability | The system always provides the advice I require to make my decision. |
| | The system performs reliably. |
| | The system responds the same way under the same conditions at different times. |
| | I can rely on the system to function properly. |
| | The system analyzes problems consistently. |
| Perceived Technical Competence | The system uses appropriate methods to reach decisions. |
| | The system has sound knowledge about this type of problem built into it. |
| | The advice the system produces is as good as that which a highly competent person could produce. |
| | The system correctly uses the information I enter. |
| | The system makes use of all the knowledge and information available to it to produce its solution to the problem. |
| Perceived Understandability | I know what will happen the next time I use the system because I understand how it behaves. |
| | I understand how the system will assist me with decisions I have to make. |
| | Although I may not know exactly how the system works, I know how to use it to make decisions about the problem. |
| | It is easy to follow what the system does. |
| | I recognize what I should do to get the advice I need from the system the next time I use it. |
| Faith | I believe advice from the system even when I don't know for certain that it is correct. |
| | When I am uncertain about a decision I believe the system rather than myself. |
| | If I am not sure about a decision, I have faith that the system will provide the best solution. |
| | When the system gives unusual advice I am confident that the advice is correct. |
| | Even if I have no reason to expect the system will be able to solve a difficult problem, I still feel certain that it will. |
| Personal Attachment | I would feel a sense of loss if the system was unavailable and I could not longer use it. |
| | I feel a sense of attachment to using the system. |
| | I find the system suitable to my style of decision making. |
| | I like using the system for decision making. |
| | I have a personal preference for making decisions with the system. |

It is noteworthy that the Scale refers to understandability but does not explicitly reference trust.

Upon close examination, it seems that the reliability factor has some redundant items. The factors titled "perceived technical competence" and "perceived understandability" might be interpreted as referencing the user's mental model of the system. For example, the item *Even if I have no reason to expect the system will be able to solve a difficult problem, I still feel certain that it will* clearly is asking about the user's mental model. Indeed, the Madsen-Gregor Scale as a whole can be understood as referring as much to evaluating the user's mental model as it does to trust. The mere fact that this distinction is fuzzy is a testament to the notion that XAI evaluation must have measures of both trust and of mental models, since the two are causally related.

One can question the appropriateness of referring to a "faith" factor. Items in this factor seem to refer to reliance and uncertainty. One can question the appropriateness of referring to a "personal attachment" factor rather than a "liking" factor.

As with other Scales, multiple interpretations are possible.  For instance, the Madsen-Gregor item *I believe advice from the system even when I don't know for certain that it is correct* asks essentially the same thing as the Cahour-Fourzy item *I am confident in the tool; it works well*.

A number of individual items are of interest, such as "It is easy to follow what the system does" and "I recognize what I should do to get the advice I need from the system." These seem to reference usability. Up to this point, issues of XAI system learnability and usability have not been considered in the XAI Program.

**Merritt Scale (2011)**

Trust is regarded as an emotional, attitudinal judgement of the degree to which the user can rely on the automated system to achieve his or her goals under conditions of uncertainty. Trust was initially broken into three factors: belief, confidence, and dependability. Factor Analysis revealed two other factors: propensity to trust and liking. The scale was evaluated in an experiment in which participants conducted a baggage screen task using a fictitious automated weapon detector in a luggage screening task. Chronbach's alpha ranged from a = .87 to a = .92.

Items in this Scale are all similar to items in the Cahour-Fourzy Scale.

1. I believe the system is a competent performer.
2. I trust the system.
3. I have confidence in the advice given by the system.
4. I can depend on the system.
5. I can rely on the system to behave in consistent ways.
6. I can rely on the system to do its best every time I take its advice.

**Schaefer Scale (2013)**

This scale was developed in the context of human-robot collaboration. Thus, trust was said to depend on both machine performance and team collaboration. Trust was analyzed into two factors: ability and performance. This scale is unique in that it is long and has a format different from all the other scales. Specifically the participant is asked to estimate the amount of time that the machine (in the study, a robot) would show each of a number of possible behaviors. In this scale format, some items are troublesome. For example, if the machine acts consistently, what is the point of asking about the percentage of time that it asks consistently? Many of the items anthropomorphize the machine (robot) and do so in ways that seem inappropriate for the XAI application (e.g., "know the difference between friend and foe," "be supportive," "be responsible," "be conscious"). For example, the point of XAI is to communicate richly and meaningfully with the participant. Thus, asking about the percentage of time that the XAI "openly communicates" or "clearly communicates" seems redundant to the evaluation of Explanation Satisfaction. In the list below, we place in the left those items that seem appropriate to XAI and in the right those that do not. The items in the left align fairly well to items in the Cahour-Fourzy Scale. One of these items "Perform a task better than a novice human user" is particularly interesting and might be added into the Cahour-Fourzy Scale.

*What percentage of the time will this machine (robot)…*

| | |
|---|---|
| Act consistently | Protect people |
| Function successfully | Act as part of the team |
| Have errors | Malfunction |
| Perform a task better than a novice human user | Clearly communicate |
| Possess adequate decision-making capability | Require frequent maintenance |
| Perform exactly as instructed | Openly communicate |
| Make sensible decisions | Know the difference between friend and foe |
| Tell the truth | Provide feedback |
| Perform many functions at one time | Warn people of potential risks in the environment |
| Follow directions | Meet the needs of the mission |
| Incompetent | Provide appropriate information |
| Dependable | Communicate with people |
| Reliable | Work best with a team |
| Predictable | Keep classified information secure |
| | Work in close proximity with people |
| | Considered part of the team |
| | Friendly |
| | Pleasant |
| | Unresponsive |
| | Autonomous |
| | Conscious |
| | Lifelike |
| | A good teammate |
| | Led astray by unexpected changes in the environment |

**Singh, et al. Scale (1993)**

This scale presupposes a context in which the participant is evaluating a device with which they have prior experience or have general familiarity with (ATMs, medical devices, etc.). Trust was defined as an attitude toward commonly encountered automated devices that reflect a potential for complacency. Trust was analyzed into five factors: confidence, reliance, trust, safety, complacency. Since the scale merges trust and reliance, it presupposes prior experience and would not be appropriate for use when a user is first learning to use an XAI. For these reasons, we feel that this scale is not appropriate for use in the XAI context. Items that might be modified to make them appropriate reference factors that are already covered in the Cahour-Fourzy Scale (i.e., trust, reliance).

Factor 1: Confidence

1. I think that automated devices used in medicine, such as CT scans and ultrasound, provide very reliable medical diagnosis.

2. Automated devices in medicine save time and money in the diagnosis and treatment of disease.

3. If I need to have a tumor in my body removed, I would choose to undergo computer-aided surgery using laser technology because it is more reliable and safer than manual surgery.

4. Automated systems used in modern aircraft, such as the automatic landing system, have made air journeys safer.

Factor 2: Reliance

1. ATMs provide a safeguard against the inappropriate use of an individual's bank account by dishonest people.

2. Automated devices used in aviation and banking have made work easier for both employees and customers.

3. Even though the automatic cruise control in my car is set at a speed below the speed limit, I worry when I pass a police radar speed trap in case the automatic control is not working properly.

Factor 3: Trust

1. Manually sorting through card catalogues is more reliable than computer-aided searches for finding items in a library.

2. I would rather purchase an item using a computer than have to deal with a sales representative on the phone because my order is more likely to be correct using the computer.

3. Bank transactions have become safer with the introduction of computer technology for the transfer of funds.

Factor 4: Safety

1. I feel safer depositing my money at an ATM than with a human teller.

2. I have to tape an important TV program for a class assignment. To ensure that the correct program is recorded, I would use the automatic programming facility on my VCR rather than manual taping.

**Wang, et al. Scale (2009)**

This scale was used to evaluate trust in a hypothetical "combat identification system" that participants used in a simulated task. All of the items were taken form or adapted from the Jian, et al. Scale. The reliability of the decisions generated by the hypothetical decision aid was a primary independent variable, in an effort to study response bias inducted by automation reliability. The scale items are reported in the paper, but not the format for the scale (e.g., was it a Likert scale?). Some items are context specific (e.g., *The aid provides security; The blue light indicates soldiers*"). What is noteworthy about some of the items is that they refer explicitly to deception and mistrust. Other items in the Wang, et al., Scale refer to trust and reliability and are covered by items in the Cahour-Fourzy Scale.

The aid is deceptive.
The aid behaves in an underhanded (concealed) manner.
I am suspicious of the aid's outputs.
I am wary of the aid.
The aid's action will have a harmful or injurious outcome.
I am confident in the aid.
The aid provides security.
The aid is dependable.
The aid is reliable.
I can trust the aid.
I am familiar with the aid.
I can trust that *blue* lights indicate soldiers.
I can trust that *red* lights indicate terrorists.

# Appendix E

# Trust Scale Recommended for XAI

This Trust Scale asks users directly whether they are confident in the XAI system, whether the XAI system is predictable, reliable, efficient, and believable.

The scale assumes that the participant has had considerable experience using the XAI system. Hence, these questions would be appropriate for scaling after a period of use, rather than immediately after an explanation has been given and prior to use experience.

A majority of the items are adapted from the Cahour-Fourzy Scale (2009), just as they have been adapted for use in other scales (e.g., Jian, et al.). In the original scale, the items are rated on a bipolar scale going from *I agree completely* to *I do not agree at all*. We have modified the items to fit the general Likert form developed for the XAI Explanation Satisfaction Scale. In addition to conforming to psychometric standards, this consistency of format will presumably make the ratings tasks easier for participants.

Item 6 was adapted from the Jian, et al. Scale, item 7 was adapted from the Schaefer Scale, and item 8 was adapted from the Madsen-Gregor Scale.

We can assert that the Recommended Scale is reliable based on these two facts:
(1). The majority of the items in this Recommended Scale essentially overlap with items in the Jian, et al. (2000) scale, which was shown empirically to be highly reliable.
(2) Items in the Recommended Scale bear overall semantic similarity to items in the Madsen-Gregor-Scale, and that scale too was also shown to have high reliability coefficients.

We can assume that the Recommended Scale has content validity Given the essential overlap of items in the Recommended Scale with items in most of the existing scales, we can safely assume that the Recommended Scale has content validity.

1. I am confident in the [tool]. I feel that it works well.

| 5 | 4 | 3 | 2 | 1 |
| --- | --- | --- | --- | --- |
| I agree strongly | I agree somewhat | I'm neutral about it | I disagree somewhat | I disagree strongly |

2. The outputs of the [tool] are very predictable.

| 5 | 4 | 3 | 2 | 1 |
| --- | --- | --- | --- | --- |
| I agree strongly | I agree somewhat | I'm neutral about it | I disagree somewhat | I disagree strongly |

3. The tool is very reliable. I can count on it to be correct all the time.

| 5 | 4 | 3 | 2 | 1 |
|---|---|---|---|---|
| I agree strongly | I agree somewhat | I'm neutral about it | I disagree somewhat | I disagree strongly |

4. I feel safe that when I rely on the [tool] I will get the right answers.

| 5 | 4 | 3 | 2 | 1 |
|---|---|---|---|---|
| I agree strongly | I agree somewhat | I'm neutral about it | I disagree somewhat | I disagree strongly |

5. The [tool] is efficient in that it works very quickly.

| 5 | 4 | 3 | 2 | 1 |
|---|---|---|---|---|
| I agree strongly | I agree somewhat | I'm neutral about it | I disagree somewhat | I disagree strongly |

6. I am wary of the [tool]. (adopted from the Jian, et al. Scale and the Wang, et al. Scale)

| 5 | 4 | 3 | 2 | 1 |
|---|---|---|---|---|
| I agree strongly | I agree somewhat | I'm neutral about it | I disagree somewhat | I disagree strongly |

7. The [tool] can perform the task better than a novice human user. (adopted from the Schaefer Scale)

| 5 | 4 | 3 | 2 | 1 |
|---|---|---|---|---|
| I agree strongly | I agree somewhat | I'm neutral about it | I disagree somewhat | I disagree strongly |

8. I like using the system for decision making.

| 5 | 4 | 3 | 2 | 1 |
|---|---|---|---|---|
| I agree strongly | I agree somewhat | I'm neutral about it | I disagree somewhat | I disagree strongly |